\documentclass[conference]{IEEEtran}
\IEEEoverridecommandlockouts

\usepackage{etoolbox}
\makeatletter
\patchcmd{\@makecaption}
  {\scshape}
  {}
  {}
  {}
\makeatletter
\patchcmd{\@makecaption}
  {\\}
  {.\ }
  {}
  {}
\makeatother

\usepackage{cite}
\usepackage{amsmath,amssymb,amsfonts}
\usepackage{algorithmic}
\usepackage{graphicx}
\usepackage{textcomp}
\usepackage{xcolor}
\def\BibTeX{{\rm B\kern-.05em{\sc i\kern-.025em b}\kern-.08em
    T\kern-.1667em\lower.7ex\hbox{E}\kern-.125emX}}

\usepackage{colortbl} 
\usepackage{pifont}
\usepackage{booktabs}
\usepackage{multirow}
\usepackage{algorithm}
\usepackage{algorithmic}
\usepackage{hyperref}
\begin{document}

\title{Data-efficient LLM Fine-tuning for Code Generation}


\author{
\IEEEauthorblockN{Weijie Lv}
\IEEEauthorblockA{
\textit{Nanjing University of}\\
\textit{Aeronautics and Astronautics}\\
Nanjing, China \\
lvweijie@nuaa.edu.cn
}
\and
\IEEEauthorblockN{Xuan Xia$^*$}
\IEEEauthorblockA{
\textit{Shenzhen Institute of Artificial}\\
\textit{Intelligence and Robotics for Society}\\
Shenzhen, China \\
xiaxuan@cuhk.edu.cn
}
\and
\IEEEauthorblockN{Sheng-Jun Huang$^*$ \thanks{$^*$ Corresponding author}}
\IEEEauthorblockA{
\textit{Nanjing University of}\\
\textit{Aeronautics and Astronautics}\\
Nanjing, China \\
huangsj@nuaa.edu.cn
}
}

\maketitle

\footnotetext[1]{Code is available at \href{https://github.com/Kyle-Lyu/data-efficient-finetuning}{https://github.com/Kyle-Lyu/data-efficient-finetuning}}

\begin{abstract}
Large language models (LLMs) have demonstrated significant potential in code generation tasks. However, there remains a performance gap between open-source and closed-source models.
To address this gap, existing approaches typically generate large amounts of synthetic data for fine-tuning, which often leads to inefficient training. 
In this work, we propose a data selection strategy in order to improve the effectiveness and efficiency of training for code-based LLMs. By prioritizing data complexity and ensuring that the sampled subset aligns with the distribution of the original dataset, our sampling strategy effectively selects high-quality data. Additionally, we optimize the tokenization process through a ``dynamic pack'' technique, which minimizes padding tokens and reduces computational resource consumption.
Experimental results show that when training on 40\% of the OSS-Instruct dataset, the DeepSeek-Coder-Base-6.7B model achieves an average performance of 66.9\%, surpassing the 66.1\% performance with the full dataset. Moreover, training time is reduced from 47 minutes to 34 minutes, and the peak GPU memory usage decreases from 61.47 GB to 42.72 GB during a single epoch. Similar improvements are observed with the CodeLlama-Python-7B model on the Evol-Instruct dataset. By optimizing both data selection and tokenization, our approach not only improves model performance but also enhances training efficiency.

\end{abstract}


\section{Introduction}

Large language models (LLMs) have recently achieved remarkable success across various domains, with their applications in code-related tasks becoming a prominent area of research. Pioneered by models such as Codex\cite{codex}, LLMs have demonstrated exceptional code processing capabilities, leading to the development of commercial products such as GitHub Copilot and open-source alternatives such as CodeLlama\cite{codellama}, DeepSeek-Coder\cite{deepseek-coder}, and StarCoder\cite{starcoder2, li2023starcoder}. However, a significant performance gap persists between open-source and closed-source models, particularly in code generation tasks.

The performance of LLMs is largely determined by the scale and quality of their training datasets, a finding underscored by recent research on scaling laws\cite{zhang2024scaling}. To improve performance, existing code LLMs\cite{codealpaca, wizardcoder, magicoder, yu2023wavecoder, song2024alchemistcoder, lei2024autocoder} have leveraged more powerful LLMs (e.g., GPT-4\cite{gpt-4}) to generate synthetic code samples for fine-tuning open-source models. Several strategies for generating code data have been proposed. Although these methods have shown promise, they often lead to inefficient optimization and training processes due to the inclusion of low-quality synthetic data within the massive training corpora.

Recent researches\cite{ouyang2022training, xia2024less, lima} have emphasized that data quality is more crucial than quantity.
LIMA\cite{lima} achieves superior performance with just 1,000 carefully curated samples. This insight raises a critical question: \textit{How can we identify the most influential training samples to enhance model performance and training efficiency simultaneously?}

Complex programming problems typically require the integration of various knowledge domains and skills, which requires more intricate reasoning processes than simpler ones. Intuitively, these complex problems may contribute substantially to model training. However, if only the complexity of the data is considered, the distribution of the sampled data may become inconsistent with that of the original dataset, negatively impacting model performance. Consequently, it is essential to consider data consistency while selecting complex training samples.

Based on these insights, we propose a data selection strategy aimed at identifying the most influential code samples. Specifically, we employ the K-Means algorithm to ensure that the distribution of the sampled data subsets reflects the original dataset. Subsequently, we compute the \textbf{I}nstruction \textbf{F}ollowing \textbf{D}ifficulty (IFD) score\cite{ifd} to quantify the complexity of each sample, selecting the most complex sample within each cluster. By selecting a smaller set of high-quality data, our approach seeks to enhance training efficiency while preserving or improving model performance.

Additionally, we introduce the ``dynamic pack'' technique for data tokenization, aiming to fully utilize the context length of LLMs and minimize the padding tokens in each batch. This is achieved by sorting the data within each batch by length, then concatenating samples in a way that ensures the total tokens do not exceed the context length of the LLM. Compared to traditional tokenization strategies, our approach not only accelerates the training process but also reduces computational resource consumption.

The experimental results demonstrate the effectiveness of our approach. When trained on 40\% of the OSS-Instruct dataset, the DeepSeek-Coder-Base-6.7B model achieves an average performance of 66.9\%, outperforming the 66.1\% achieved on the full dataset. Moreover, the training time is reduced from 47 minutes to 34 minutes, and the peak GPU memory usage decreases from 61.47 GB to 42.72 GB during one training epoch when employing our approach. These improvements are consistently observed when applying our approach to the CodeLlama-Python-7B model on the Evol-Instruct dataset.

The main contributions of this paper are as follows:
\begin{itemize}
\item We propose a data selection strategy for code samples that identifies influential training data by considering both complexity and consistency.
\item We introduce the ``dynamic pack'' technique during data tokenization, which significantly reduces padding tokens and improves training efficiency and resource utilization.
\item Extensive experiments validate the effectiveness of our approach, demonstrating superior performance with less data and significantly improving training efficiency.
\end{itemize}

\section{Related Work}

\subsection{Code LLMs}
The advancement of LLMs has significantly impacted various domains, including code generation and understanding. Closed-source models, such as GPT-4\cite{gpt-4}, have consistently ranked highly in the mainstream evaluation metrics, demonstrating superior performance in code-related tasks. These models leverage substantial computational resources and proprietary datasets, resulting in a performance gap between them and their open-source counterparts.

To bridge this gap and democratize access to advanced coding capabilities, several open-source models have been developed. Notable among these are CodeLlama\cite{codellama}, DeepSeek-Coder\cite{deepseek-coder}, and CodeGemma\cite{codegemma}. CodeLlama, derived from the Llama 2\cite{llama2} architecture, is specifically fine-tuned for code generation tasks and has shown competitive results compared to closed-source models. These open-source models have significantly advanced the field of code generation. They offer robust alternatives to closed-source models, promoting innovation and collaboration within the research community. 

\subsection{Data Generation}
A significant area of research focuses on generating instruction data to fine-tune LLMs. Self-Instruct\cite{self-instruct} is one such method that refines weaker student models by using strong teacher models to generate synthetic instructions. This approach leverages the expertise of advanced LLMs to produce diverse and complex instruction data, which in turn helps to train more effective student models. Evol-Instruct\cite{wizardcoder} extends this by iteratively increasing the complexity of the instructions. This method evolves the instructions to be more challenging over multiple iterations, thereby improving the model's capacity to handle complex tasks. Another notable approach is OSS-Instruct\cite{magicoder}, which generates realistic coding problems based on open-source code snippets. By extracting real-world code segments from repositories such as GitHub, OSS-Instruct prompts LLMs to create relevant and practical coding challenges, ensuring the generated data closely mirrors actual programming scenarios.

These methodologies typically utilize more powerful LLMs, such as GPT-4, to generate large volumes of synthetic data. These approaches address the issue of data quantity, but often overlook data quality. Ensuring the relevance, diversity, and complexity of the generated data remains a critical challenge for further improving model performance.

\subsection{Automated Data Selection}

The process of manual data curation is not only costly but also prone to subjective bias, making the development of automated data selection methods critically important. Current automated data selection methods are broadly categorized into two types: external model-assisted \cite{alpagasus, instag, lift, bukharin2023data, wang2024inscl} and non-external model-assisted \cite{ifd, dq, zhang2024recost, he2024shed}.

For external model-assisted methods, AlpaGasus\cite{alpagasus} utilizes carefully crafted prompt templates to leverage ChatGPT for data quality scoring. InsTag\cite{instag} employs ChatGPT to generate detailed labels for each data instruction, evaluating the complexity and diversity of the data based on these labels. LIFT\cite{lift} generates a diverse set of instructions using GPT-4 to augment the dataset, followed by vectorization and selection of subsets based on row variables, ultimately using GPT-4 for multi-dimensional scoring of the data. These methods have demonstrated efficacy in handling large-scale datasets, but incur significant economic costs.

For non-external model-assisted methods, DQ\cite{dq} integrate techniques of data distillation and coreset selection \cite{iyer2021submodular}. The core technology involves defining a gain function to iteratively partition the dataset and select representative samples, thereby maximizing the diversity of data. Cherry LLM \cite{ifd} introduces the \textbf{I}nstruction \textbf{F}ollowing \textbf{D}ifficulty (IFD) score, determined by comparing the cross-entropy loss of model-generated responses with and without instructions. A high IFD score implies that the model struggles to accurately align answers with instructions, reflecting the complexity of the samples.

Despite the progress made in automated data curation, methods specifically designed for code data selection remain notably underexplored in the literature.

\section{Methodology}

\begin{figure*}
    \centering
    \includegraphics[width=0.7\textwidth]{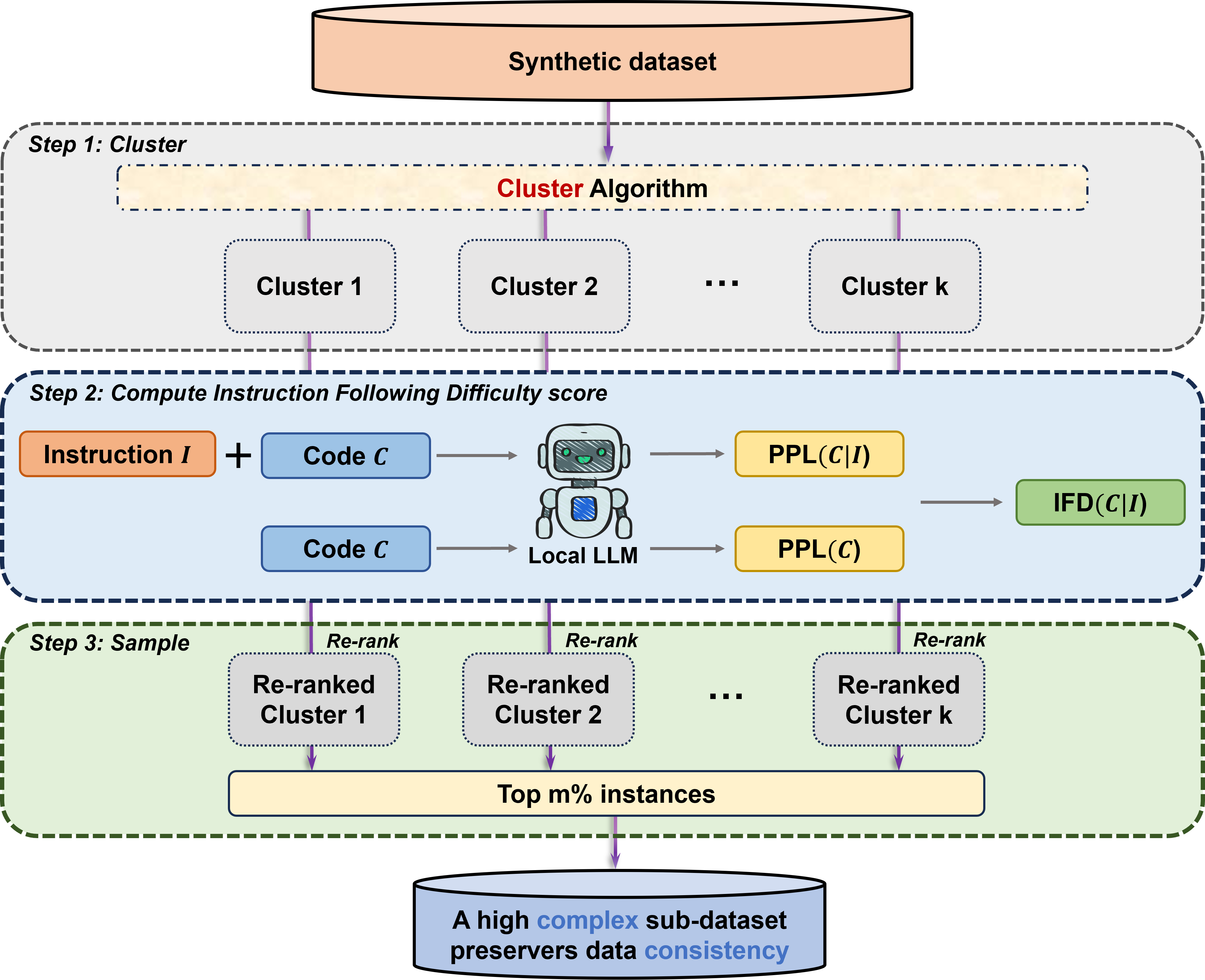}
    \caption{The overview of our proposed data selection strategy, including three steps. Step 1: Partitioning the synthetic dataset into multiple clusters. Step 2: Computing the Instruction Following Difficulty score by comparing the model's perplexity with and without instructions. Step 3: Sampling the top m\% instances from each re-ranked cluster to form a high-complexity sub-dataset that preserves data consistency. Finally, the selected data is used for fine-tuning open-source code LLMs.}
    \label{fig:data-selection}
\end{figure*}

\subsection{Task Definition}

Let $D$ denote a dataset consisting of $n$ pairs $x = (Instruction, Code)$, representing instruction tuning data samples. Given a large pool of instruction tuning dataset $D = \{x_1, x_2, \ldots, x_n\}$, where each $x_i$ represents an individual instruction-code pair ($I_i$, $C_i$), our objective is to select a subset $S_\pi^{(m)}$ of size $m$ from $D$, utilizing a selection strategy $\pi$. To evaluate the effectiveness of the selected subset, we represent the model performance after instruction tuning as $P$. The optimal selection strategy $\pi^*$  within a given data budget $m$ is then defined as:
\begin{equation}
\pi^* = \arg\max_\pi P(S_\pi^{(m)}).
\label{eq:1}
\end{equation}

\subsection{Data Selection Strategy}

The objective of our data selection strategy is to curate a high-quality and representative subset of data, ensuring that training on the selected samples yields performance comparable to or better than training on the entire dataset. Specifically, we focus on two critical aspects: complexity and consistency. Complexity ensures that the selected data contain challenging coding tasks, while consistency ensures that the subset preserves the distributional properties of the original dataset. The overview of our data selection strategy is illustrated in Fig. \ref{fig:data-selection}.

The quality of a dataset is often determined by its ability to represent a wide range of problem-solving scenarios. In coding tasks, complex problems typically require the integration of multiple domains of knowledge and advanced reasoning skills, making them more valuable for model training. However, selecting only complex samples without considering the overall distribution of the dataset may lead to biased or incomplete learning. Therefore, our strategy balances complexity and consistency to ensure that the selected data are both diverse and representative.

To measure the complexity of a coding sample, we utilize the \textbf{I}nstruction \textbf{F}ollowing \textbf{D}ifficulty (IFD) score\cite{ifd}. The IFD score quantifies the difficulty of generating an appropriate response to an instruction by comparing the model's perplexity with and without the instruction. For a given sample \( (I_i, C_i) \), where $I_i$ represents the instruction and $C_i$ represents the corresponding code, the IFD score is defined as:

\begin{equation}
    \text{PPL}(C_i \mid I_i) = e^{ -\frac{1}{N} \sum_{j=1}^{N} \log P(C_{i,j} \mid I_i, C_{i,1}, \ldots, C_{i,j-1}) },
\label{eq:2}
\end{equation}

\begin{equation}
    \text{PPL}(C_i ) = e^{ -\frac{1}{N} \sum_{j=1}^{N} \log P(C_{i,j} \mid C_{i,1}, \ldots, C_{i,j-1}) },
\label{eq:3}
\end{equation}

\begin{equation}
    \text{IFD}(C_i \mid I_i) = \frac{\text{PPL}(C_i \mid I_i)}{\text{PPL}(C_i)},
\label{eq:4}
\end{equation}


where \( N \) represents the length of the code response \( C_i \), and \( C_{i,j} \) denotes the \( j \)-th token in \( C_i \). A higher IFD score indicates that the sample likely involves more complex knowledge or rare combinations of concepts, suggesting a more challenging problem. In contrast, a lower IFD score suggests that the sample is relatively simpler, containing information that aligns with the model's pre-trained knowledge.

To ensure the consistency of the selected subset with the original dataset, we employ the K-Means algorithm to partition the dataset into distinct clusters. This approach ensures that the selected subset preserves the distributional properties of the original dataset.

Our approach is summarized as follows. 
First, we utilize lightweight sentence transformers\cite{reimers-2019-sentence-bert} to derive embeddings from the given instructions. These embeddings effectively capture the semantic essence of the instructions, facilitating more accurate clustering. Next, we apply the K-Means algorithm to partition the dataset into $k$ clusters. Within each cluster, we compute the IFD score for each sample, then reorder the data based on these scores. Finally, we sample the top $m\%$ of samples from each re-ranked cluster. This sampling strategy ensures a balance between complexity and consistency, enhancing the model's ability to generalize across diverse coding tasks.

\subsection{Data Tokenization Strategy}

Tokenization is a crucial step in the fine-tuning processes of LLMs. Its primary objective is to segment text into smaller units for processing. However, due to the variability in sample lengths, traditional strategies typically align samples to the model's maximum input length or to the longest sequence within a batch by adding padding tokens. This often results in a high proportion of padding tokens. Although these padding tokens do not contribute to the model's forward computation, they consume computational resources and reduce training efficiency.

To address this issue, we propose the ``dynamic pack'' technique for data tokenization, inspired by the ``pack'' technique in T5\cite{raffel2020exploring}. The ``pack'' technique truncates samples to a maximum length of 512 tokens before concatenating multiple samples to ensure the total tokens are consistent across batches. In contrast, our ``dynamic pack'' technique aims to fully utilize the context length of LLMs while minimizing the number of padding tokens. This method first sorts the samples within a batch by length, without truncating any of them, and then attempts to concatenate multiple samples in a way that ensures the total length does not exceed the model's context length. This process generates a new batch, which is subsequently padded to the maximum length of the concatenated samples.
This approach significantly reduces the proportion of padding tokens and improves training efficiency, as demonstrated by the experimental results in \textit{Section \ref{subsec:Training efficiency}}.

\section{Experiments}

\subsection{Datasets}

\textbf{EVOL-Instruct.} The EVOL-Instruct\cite{wizardcoder} dataset is derived from the iterative evolution of the Code Alpaca\cite{codealpaca} dataset, where the complexity of instructions is incrementally improved using ChatGPT with evolution prompts. These prompts encompass five distinct aspects, including the imposition of constraints, the substitution of broad requirements with more detailed ones, the extension of reasoning steps, the inclusion of deceptive code, and the enhancement of time or space complexity. Each instruction undergoes multiple iterations of evolution, during which pruning and post-processing are conducted to eliminate undesirable instructions and responses. This method of iteratively increasing complexity yields instructions of higher quality and depth. We utilize the \textit{Evol-Instruct-Code-80K\footnote{https://huggingface.co/datasets/nickrosh/Evol-Instruct-Code-80k-v1}} dataset, an open-source implementation comprising approximately 78K samples.

\textbf{OSS-Instruct.} The \textit{OSS-Instruct}\footnote{https://huggingface.co/datasets/ise-uiuc/Magicoder-OSS-Instruct-75K} dataset utilizes ChatGPT to generate programming problems and their corresponding solutions. The dataset's uniqueness stems from its use of authentic code snippets from open-source repositories, such as GitHub, as seeds to guide the generation. This method is particularly effective as it draws inspiration from real-world coding examples, encouraging the language model to produce problems that mirror actual programming challenges. This approach not only ensures the diversity and authenticity of the generated samples, but also captures the various challenges encountered in real-world programming.

\subsection{Benchmarks}

We employ four code benchmarks: HumanEval\cite{humaneval}, HumanEval+\cite{evalplus}, MBPP\cite{mbpp}, and MBPP+\cite{evalplus}. Consistent with previous research\cite{magicoder, evalplus, chen2024teaching}, we use greedy decoding to generate a single code sample for each benchmark and LLM, focusing our comparison on the pass@1 metric.

\begin{table*}[ht]
\renewcommand{\arraystretch}{1.2}
\centering
\caption{Performance comparison of different models using greedy decoding, demonstrating that our approach significantly improves training efficiency while maintaining or even improving model performance. The results for Magicoder-DS are sourced from their paper\cite{magicoder}, and the results for WizardCoder-Python are sourced from EvalPlus\cite{evalplus} leaderboard. The remaining results are obtained through our experimental evaluations.}
\begin{tabular}{lccccccc}
\toprule
\multirow{2}{*}{\textbf{Model}} & \multirow{2}{*}{\textbf{Size}} & \multirow{2}{*}{\textbf{Training Data}} & \multicolumn{5}{c}{\textbf{Benchmark(Pass@1 \%)}} \\
\cmidrule(lr){4-8}
& & & \textbf{HumanEval} & \textbf{HumanEval+} & \textbf{MBPP} & \textbf{MBPP+} & \textbf{Average} \\
\midrule
DeepSeek-Coder-Base & 6.7B & - & 47.6 & 40.2 & 69.2 & 54.6 & 52.9 \\
CodeLlama-Python & 7B & - & 39.0 & 34.1 & 58.1 & 46.1 & 44.3 \\
\midrule
\multicolumn{8}{c}{\textit{Models trained on OSS-Instruct dataset}} \\
\midrule
Magicoder-DS & 6.7B & 75K & 66.5 & 60.4 & 75.4 & 61.9 & 66.1 \\
Ours-DS (full data) & 6.7B & 75K & 65.2 & 61.0 & \textbf{75.9} & \textbf{62.4} & 66.1 \\
Ours-DS (40\%) & 6.7B & 30K & \textbf{68.3} & \textbf{61.6} & \textbf{75.9} & 61.7 & \textbf{66.9} \\
Ours-DS (30\%) & 6.7B & 23K & 65.9 & 59.1 & 75.4 & 62.2 & 65.7 \\
\cmidrule(lr){1-8}
Ours-CL (full data) & 7B & 75K & 51.8 & 46.3 & \textbf{61.2} & \textbf{50.9} & 52.6 \\
Ours-CL (40\%) & 7B & 30K & 54.3 & 50.0 & 60.4 & 50.4 & \textbf{53.8} \\
Ours-CL (30\%) & 7B & 23K & \textbf{55.5} & \textbf{50.6} & 59.9 & 48.1 & 53.5 \\
\midrule
\multicolumn{8}{c}{\textit{Models trained on EVOL-Instruct dataset}} \\
\midrule
Ours-DS (full data) & 6.7B & 78K & \textbf{67.7} & 60.4 & 71.4 & 58.1 & 64.4 \\
Ours-DS (40\%) & 6.7B & 31K & 67.1 & \textbf{61.0} & \textbf{72.4} & \textbf{59.1} & \textbf{64.9} \\
Ours-DS (30\%) & 6.7B & 23K & \textbf{67.7} & 59.1 & 69.9 & 58.9 & 63.9 \\
\cmidrule(lr){1-8}
WizardCoder-Python & 7B & 78K & 50.6 & 45.1 & 58.5 & 49.5 & 50.9 \\
Ours-CL (full data) & 7B & 78K & 53.3 & \textbf{49.0} & 60.7 & 48.6 & \textbf{52.9} \\
Ours-CL (40\%) & 7B & 31K & 53.0 & 47.0 & 60.7 & \textbf{49.9} & 52.7 \\
Ours-CL (30\%) & 7B & 23K & \textbf{53.7} & 47.6 & \textbf{61.7} & 48.6 & \textbf{52.9} \\
\bottomrule
\end{tabular}
\label{tab:main_results}
\end{table*}

\textbf{HumanEval/HumanEval+.} HumanEval and its enhanced counterpart, HumanEval+, serve as critical benchmarks for evaluating the code generation capabilities of LLMs. HumanEval comprises 164 manually-written Python problems, with an average of 9.6 tests per problem. HumanEval+ extends this benchmark by significantly increasing the number of test cases through automated test input generation techniques, including the use of LLMs and mutation strategies, resulting in a more rigorous evaluation framework.

\textbf{MBPP/MBPP+.} The MBPP (Mostly Basic Python Programming) benchmark includes approximately 1,000 Python challenges, crowd-sourced to assess fundamental programming skills and standard library usage. These challenges are geared towards beginners and each provides a description, a solution, and three tests to verify solution accuracy. MBPP+ enhances the MBPP benchmark by incorporating a subset of hand-verified problems from MBPP-sanitized dataset, ensuring that the tasks are well-defined and unambiguous. This extension enhances the benchmark's reliability and applicability in more rigorous evaluations.

\subsection{Implementation Details}

In our experiments, we refer to the experimental results in DQ \cite{dq} and partition the dataset into 10 clusters using K-Means algorithm. We employ two base models for our experiments, including DeepSeek-Coder-Base-6.7B and CodeLlama-Python-7B. Our models are fine-tuned for 3 epochs using four NVIDIA A100-80GB GPUs with the Fully Sharded Data Parallel (FSDP) module in PyTorch. Specifically, we use AdamW\cite{adamw} as our optimizer with a learning rate of 5e-5, a cosine learning rate scheduler, and 15 warmup steps. For comparative analysis of tokenization strategies, the maximum sequence length is set to 4096. The global batch size for all experiments is set to 256. 

\subsection{Main Results}

Here, we highlight the performance improvements of our selection method compared to full dataset training. Table \ref{tab:main_results} presents the results of different models on the OSS-Instruct and Evol-Instruct datasets, evaluated on the HumanEval(+) and MBPP(+) benchmarks with greedy decoding.

On the OSS-Instruct dataset, the DeepSeek-Coder-Base-6.7B (DS-Base-6.7B) and CodeLlama-Python-7B (CL-Python-7B) models, when trained with our proposed data selection strategy, maintain or even improve performance while reducing the amount of training data. 
Specifically, the DS-Base-6.7B model, trained on 40\% of the data, achieves the highest performance on most benchmarks, with an average score of 66.9\%. This surpasses the score of 66.1\% obtained when trained on the full dataset. 
Similarly, The CL-Python-7B model, trained on 40\% of the data, outperforms the model trained with the whole dataset on the HumanEval(+) benchmarks and achieves comparable performance on the MBPP(+) benchmarks. It yields an average score of 53.8\%, which is higher than the 52.6\% achieved with the full dataset.
These results demonstrate that our selection strategy significantly improves training efficiency while maintaining or even improving model performance.

Experiments on the Evol-Instruct dataset further validate the effectiveness of our data selection strategy. 
For the DS-Base-6.7B model, training on 40\% of the data results in performance that is either better than or comparable to that achieved with the full dataset. It achieves an average score of 64.9\%, surpassing the 64.4\% obtained with the full dataset.
For the CL-Python-7B model, training on 30\% of the data yields the best performance on both HumanEval and MBPP benchmarks, with an average score of 52.9\%. This matches the performance of the model trained on the full dataset.

Overall, these results indicate that model performance is not solely dependent on the quantity of data but is significantly influenced by data quality. 
The OSS-Instruct dataset exhibits higher overall quality compared to the Evol-Instruct dataset. This discrepancy may be attributed to the different methodologies used in their generation processes. Compared to the OSS-Instruct dataset which utilizes diverse code snippets to guide the language model in generating samples, the Evol-Instruct dataset generates data by progressively increasing the complexity of existing instructions, resulting in more homogeneous data.
By prioritizing high-complexity subsets while ensuring that the distribution aligns with the original dataset, our data selection strategy demonstrates its effectiveness across different models and datasets. Selecting high-quality data not only improves performance but also significantly improves training efficiency.

\begin{figure*}[t]
    \centering
    \includegraphics[width=0.9\textwidth]{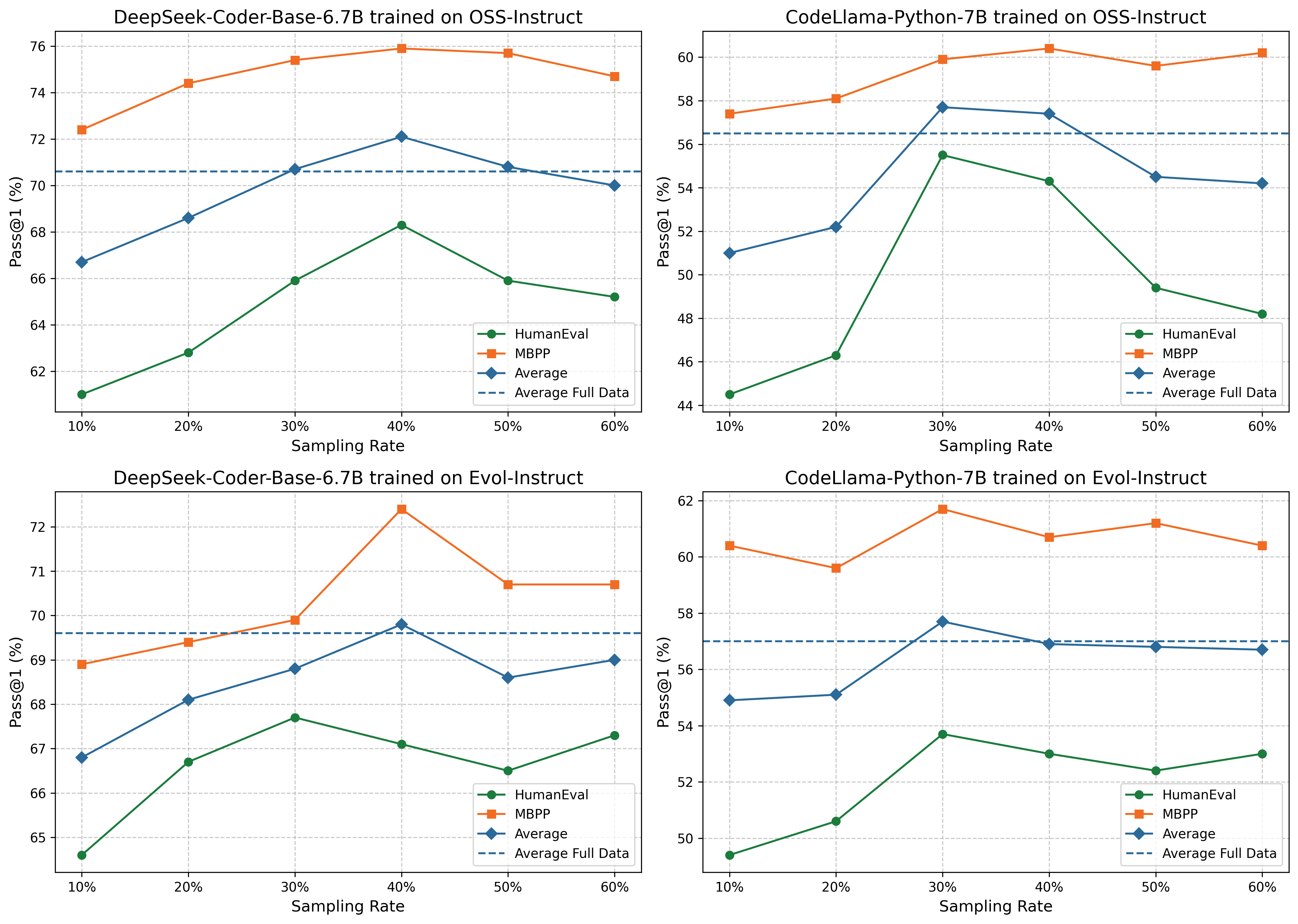}
    \caption{Impact of sampling rates on model performance. The results demonstrate that model performance peaks at sampling rates between 30\% and 40\%, after which it begins to decline. ``Average Full Data'' denotes the model's average performance on the HumanEval and MBPP benchmarks when trained on the full data. ``Average'' reflects the model's average performance on these two benchmarks at different sampling rates.}
    \label{fig:sr_comparison}
\end{figure*}

\section{Ablation Studies}



\subsection{Sampling Rates}

To examine the impact of sampling rates on model performance, we present the results using the DeepSeek-Coder-Base-6.7B (DS-Base-6.7B) and CodeLlama-Python-7B (CL-Python-7B) models on the OSS-Instruct and Evol-Instruct datasets. The results, shown in Fig. \ref{fig:sr_comparison}, compare the models' performance at different sampling rates (ranging from 10\% to 60\%) against the performance based on the full dataset.

The results show that model performance does not exhibit a linear correlation with the sampling rates. The performance peaks at sampling rates between 30\% and 40\%, after which it gradually declines.
This phenomenon reflects the presence of noise and redundant data in synthetic datasets: as the sampling rate increases, the incorporation of lower-quality data samples can degrade model performance. By prioritizing data complexity and aligning the sampled subset with the original dataset distribution, our strategy effectively selects high-quality data. 
For the OSS-Instruct dataset, the DS-Base-6.7B model achieves its best average performance (72.1\%) at a sampling rate of 40\%, outperforming the result obtained with the full dataset (70.6\%). Similarly, the CL-Python-7B model achieves its best average performance (57.7\%) at a sampling rate of 30\%, also surpassing that with the full dataset (56.5\%). For the Evol-Instruct dataset, the DS-Base-6.7B model achieves its best average performance (69.8\%) at a sampling rate of 40\%, slightly higher than the performance with the full dataset (69.6\%). The CL-Python-7B model achieves its best average performance (57.7\%) at a sampling rate of 30\%, outperforming the full dataset performance (57.0\%).

These results indicate that the optimal sampling rate for the DS-Base-6.7B model is 40\%, while for the CL-Python-7B model, it is 30\%. This discrepancy is likely associated with the models' learning capabilities. As shown in Table \ref{tab:main_results}, DS-Base-6.7B significantly outperforms CL-Python-7B on all benchmarks, suggesting that DS-Base-6.7B exhibits superior generalization capabilities. Consequently, the DS-Base-6.7B model benefits from a higher sampling rate, enabling it to leverage more data. Furthermore, we hypothesize that different synthetic datasets may also contribute to the variations in model performance at different sampling rates.

\begin{table}[t]
\renewcommand{\arraystretch}{1.2}
\centering
\caption{Performance comparison with different data selection strategies. DS denotes DeepSeek-Coder, and CL denotes CodeLlama.}
\begin{tabular}{lclcc}
\toprule
\textbf{Model} & \textbf{Sample Rate} & \textbf{Strategy} & \textbf{HumanEval} & \textbf{MBPP}\\ 
\midrule
\multicolumn{5}{c}{\textit{OSS-Instruct dataset}} \\
\midrule
\multirow{4}{*}{DS-Base-6.7B} & \multirow{4}{*}{40\%} & Random & 62.2\% & 75.7\% \\
& & IFD & 62.8\% & 74.7\% \\
& & K-Means & 62.2\% & 72.9\% \\
& & Ours & \textbf{68.3\%} & \textbf{75.9\%} \\
\midrule
\multicolumn{5}{c}{\textit{Evol-Instruct dataset}} \\
\midrule
\multirow{4}{*}{CL-Python-7B} & \multirow{4}{*}{30\%} & Random & 49.4\% & 60.2\% \\
& & IFD & 53.0\% & 59.4\% \\
& & K-Means & 51.2\% & 59.9\% \\
& & Ours & \textbf{53.7\%} & \textbf{61.7\%} \\
\bottomrule
\end{tabular}
\label{tab:sample_strategies_comparison}
\end{table}

\subsection{Data Selection Strategies}

Through experimental analysis of different data selection strategies on the OSS-Instruct and Evol-Instruct datasets shown in Table \ref{tab:sample_strategies_comparison} , we observe that data selection strategies significantly impact the model performance.

On the OSS-Instruct dataset, the DS-Base-6.7B model, with a sampling rate of 40\%, achieves scores of 62.2\% and 75.7\% on the HumanEval and MBPP benchmarks, respectively, when employing a random selection strategy. The IFD-based strategy, which prioritizes complex samples, improves the HumanEval score to 62.8\% but reduces the MBPP score to 74.7\%. This suggests that selecting only high-complexity samples may introduce data distribution bias, thereby negatively affecting the model's performance. To address this, we incorporate the K-Means algorithm to maintain consistency in data distribution, setting the number of clusters to 10 based on the experimental results in DQ\cite{dq}. Our proposed strategy achieves the best performance, with scores of 68.3\% and 75.9\% on the HumanEval and MBPP benchmarks, respectively, significantly outperforming other strategies.

A similar trend is observed on the Evol-Instruct dataset with the CL-Python-7B model at a sampling rate of 30\%. The random selection strategy yields scores of 49.4\% on HumanEval and 60.2\% on MBPP, while the IFD strategy improves the HumanEval score to 53.0\% but slightly reduces the MBPP score to 59.4\%. The K-Means strategy achieves scores of 51.2\% on HumanEval and 59.9\% on MBPP, whereas our proposed strategy delivers the best performance, with scores of 53.7\% on HumanEval and 61.7\% on MBPP.

These results highlight that an effective data selection strategy should balance sample complexity and data distribution, as relying solely on complexity-based selection or random sampling does not lead to optimal performance.

\subsection{Training efficiency}
\label{subsec:Training efficiency}

In Table \ref{tab:efficiency_comparison}, we evaluate the effectiveness of our ``dynamic pack'' technique in optimizing training efficiency across different models and datasets. The results include the training time and the peak GPU memory usage during one training epoch. To monitor the GPU memory consumption, we utilize the \texttt{torch.cuda.max\_memory\_allocated} function.

\begin{table}[t]
\renewcommand{\arraystretch}{1.2}
\centering
\caption{Comparison of training efficiency with different tokenization strategies for training one single epoch. DS denotes DeepSeek-Coder, and CL denotes CodeLlama.}
\begin{tabular}{cccc}
\toprule
\multirow{2}{*}{\textbf{Strategy}} & \textbf{Padding} & \multicolumn{2}{c}{\textbf{Training}} \\
\cmidrule(lr){3-4}
 & \textbf{Rate} & \textbf{Time} & \textbf{Peak Memory} \\
\midrule
\multicolumn{4}{c}{\textit{DS-Base-6.7B trained on OSS-Instruct dataset}} \\
\midrule
 w/o dynamic pack & 36.54\% & 47 min & 61.47 GB \\
 w dynamic pack & 15.24\% & 34 min & 42.72 GB \\
\midrule
\multicolumn{4}{c}{\textit{CL-Python-7B trained on Evol-Instruct dataset}} \\
\midrule
 w/o dynamic pack & 54.41\% & 56 min & 60.08 GB \\
 w dynamic pack & 17.23\% & 28 min & 47.40 GB \\
\bottomrule
\end{tabular}
\label{tab:efficiency_comparison}
\end{table}

The results demonstrate that the ``dynamic pack'' technique significantly reduces the padding token rate. 
Specifically, for the DS-Base-6.7B model, the padding rate decreases from 36.54\% to 15.24\%, while for the CL-Python-7B model, it decreases from 54.41\% to 17.23\%. 
This substantial reduction is primarily attributed to the technique's dynamic concatenation of multiple samples, which maximizes the utilization of the model's context length and minimizes the number of padding tokens. In contrast, traditional padding strategies, which align samples to either the model's maximum input length or the longest sequence within a batch, introduce a large amount of padding tokens.

In addition to reducing padding tokens, our ``dynamic pack'' technique also effectively improves training efficiency by decreasing both training time and peak GPU memory usage. 
For the DS-Base-6.7B model, the training time decreases from 47 minutes to 34 minutes, and the peak GPU memory usage decreases from 61.47 GB to 42.72 GB. Similarly, for the CL-Python-7B model, the training time decreases from 56 minutes to 28 minutes, and the peak GPU memory usage decreases from 60.08 GB to 47.40 GB.
These improvements are primarily attributed to the reduction in padding tokens, which lowers computational overhead and alleviates memory pressure. 

In summary, our ``dynamic pack'' technique, through dynamic sample concatenation, significantly reduces padding rates, shortens training time, and lowers peak memory usage. These results demonstrate its efficiency and practicality in instruction fine-tuning tasks.

\section{Conclusion}

In this work, we propose a data selection and tokenization strategy aimed at improving the performance and efficiency of open-source LLMs for code generation. By utilizing the K-Means algorithm and the IFD score, we select high-quality, complex training samples while maintaining distribution consistency with the original dataset. Additionally, we introduce the ``dynamic pack'' technique, which minimizes padding tokens and significantly improves training efficiency.

The results across different models and datasets demonstrate the effectiveness of our proposed method. The model achieves competitive or superior performance with reduced data, while both training time and peak GPU memory are significantly decreased. This substantially improves training efficiency.

Future work should focus on ensuring the correctness of complex code data. Although our data selection strategy effectively selects influential data, it will be essential to incorporate mechanisms for validating and refining complex code in order to further improve model performance.

\section*{Acknowledgment}

We would like to express our sincere gratitude to those who have contributed to this work. Additionally, we utilize ChatGPT to review and refine sentence structures, which helped enhance the clarity and quality of the text.

This research was funded by Guangdong Basic and Applied Basic Research Foundation (2024A1515012026, 2021A1515110700, 2023A1515012570), National Natural Science Foundation of China (62106155), Longang District Shenzhen's ``Ten Action Plan'' (LGKCSDPT2024002, LGKCSDPT2024003), and Shenzhen Institute of Artificial Intelligence and Robotics for Society (AIRS).

\bibliography{ref}
\bibliographystyle{IEEEtran}

\end{document}